\documentclass[letterpaper, 10 pt, conference]{ieeeconf}  % Comment this line out if you need a4paper

\IEEEoverridecommandlockouts                              % This command is only needed if 
\usepackage{bm}
\usepackage{graphicx}
\usepackage{amsmath}
\usepackage{amsfonts}
\usepackage{amssymb}
\usepackage{multirow}
\usepackage{wrapfig}
\usepackage{color}

\usepackage{cite}    
\makeatletter
\let\NAT@parse\undefined
\makeatother

\usepackage{algorithm}
\usepackage{algorithmic}
\usepackage{booktabs}
\usepackage{afterpage}
\usepackage{hyperref}

\newtheorem{theorem}{Theorem}

\title{\LARGE \bf
CARO: Contact-Agnostic Residual Observation for Zero-Shot Robust Quadruped Locomotion
}

\author{Zihan Yang$^{1}$, Shixuan Han$^{2}$, Kexin Guo$^{1,\dagger}$, and Xiang Yu$^{2}$% <-this % stops a space
\thanks{$\dagger$ Corresponding author.}
\thanks{$^{1}$School of Aeronautic Science and Engineering, Beihang University, Beijing 100035, China}%
\thanks{$^{2}$School of Automation Science and Electrical Engineering, Beihang University, Beijing 100035, China}%
}

\begin{document}

\maketitle
\thispagestyle{empty}
\pagestyle{empty}

%%%%%%%%%%%%%%%%%%%%%%%%%%%%%%%%%%%%%%%%%%%%%%%%%%%%%%%%%%%%%%%%%%%%%%%%%%%%%%%%
\begin{abstract}
We propose CARO, a contact-agnostic residual observation framework for policy adaptation. CARO embeds a fixed-base Euler--Lagrange model into the reinforcement learning control loop and constructs a torque-level residual observation without requiring torque sensors, explicit contact estimation, or vision-based measurements of the floating-base position and linear velocity. A disturbance observer extracts a structured signal representing dynamics mismatch, while the policy learns to exploit this feedback for online adaptation. CARO is trained under the same terrain, command, and domain-randomization conditions as the nominal policy, without specialized disturbance curricula or additional adaptation supervision. Nevertheless, it achieves substantially improved zero-shot robustness in simulation and sim-to-real transfer tasks involving out-of-distribution payloads, center-of-mass shifts, terrain geometries, abrupt dynamics changes, and elevated-platform landings.
\end{abstract}

%%%%%%%%%%%%%%%%%%%%%%%%%%%%%%%%%%%%%%%%%%%%%%%%%%%%%%%%%%%%%%%%%%%%%%%%%%%%%%%%
\section{Introduction}

Reinforcement learning (RL) has enabled agile and versatile quadruped locomotion, achieving impressive performance in high-speed running \cite{he_agile_2024}, perturbation recovery \cite{DeepDOB_locomotion}, and complex-terrain traversal \cite{kumar_rma_2021}. Despite this progress, deployment robustness remains a persistent limitation. When a robot encounters conditions outside the training distribution, the performance of learned policies can degrade significantly. Most existing pipelines achieve robustness implicitly, either through large-scale simulation and domain randomization or through adaptation modules that infer latent environment properties and adjust policy behavior accordingly \cite{kumar_rma_2021,nahrendra_dreamwaq_2023,xiao_safe_2024,long_hybrid_2024}. Although effective under moderate distribution shifts, these approaches generally lack an explicit, physically structured feedback channel for detecting abrupt changes in the robot's dynamics and contact conditions.

A complementary line of research incorporates model-based control structures into learning-based policies \cite{ji_concurrent_2022,huang_datt_2023,lyu_rl2ac_2024,gao_neural_2025}. Predictive errors, disturbance estimates, and adaptive signals can provide informative closed-loop feedback to the policy and improve robustness under model mismatch. However, applying internal-model-based and observer-based methods to quadruped locomotion remains challenging because legged dynamics are hybrid, nonlinear, and strongly affected by contact interactions. Accurate floating-base inverse-dynamics calculations typically require reliable estimates of the floating-base state, actuator behavior, and contact forces. These sensing and modeling requirements can be difficult to satisfy during terrain transitions, foot impacts, slipping, and large body motions, limiting the direct deployment of conventional observer-based feedback in learned locomotion systems.

\begin{figure}[t!]
    \centering
    \includegraphics[width=\linewidth]{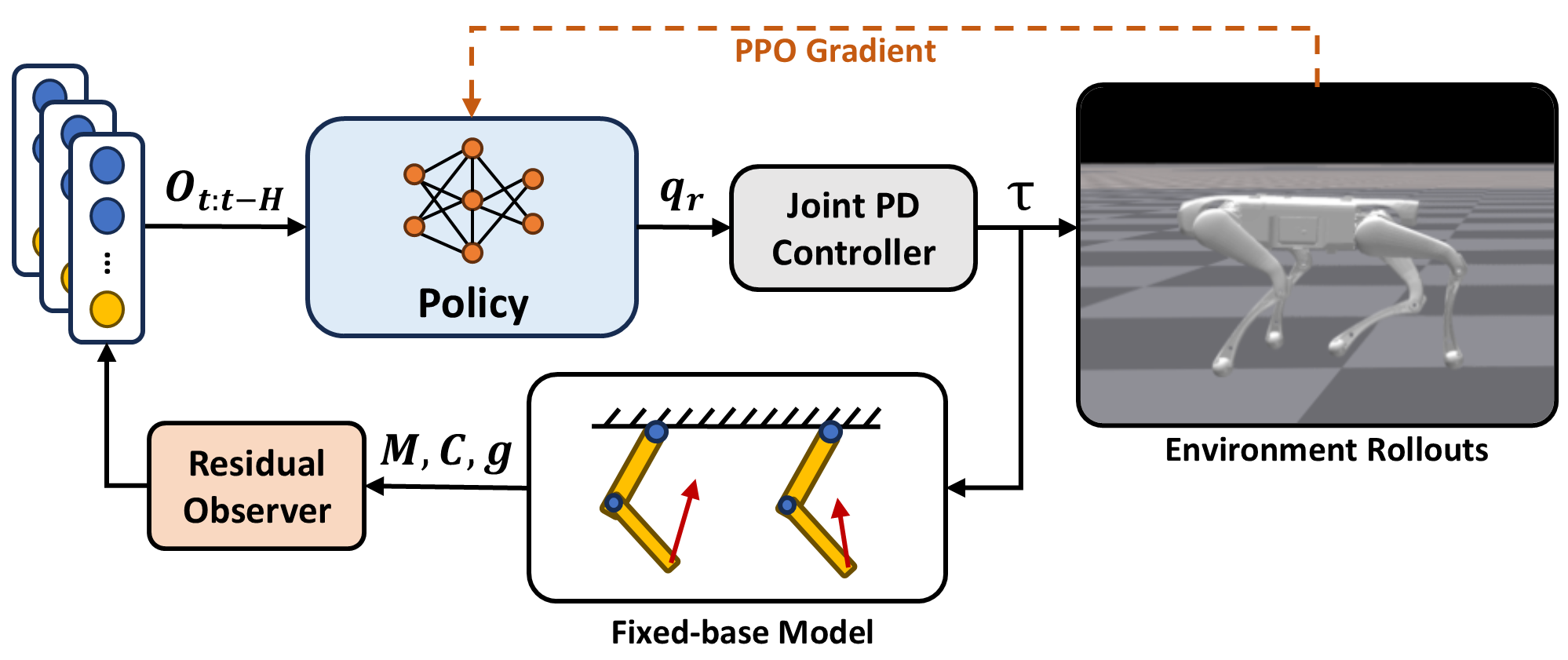}
    \caption{Overview of the CARO framework. The fixed-base Euler--Lagrange internal model is embedded within the RL control loop, and the resulting joint-level residual is provided to the policy as an adaptation signal.}
    \label{fig:caro_framework}
\end{figure}

In this work, we propose Contact-Agnostic Residual Observation (CARO), a
framework for robust quadruped locomotion. CARO embeds a
fixed-base Euler--Lagrange internal model into the RL control loop and
constructs a lightweight joint-level residual observation. Although the
internal model omits explicit floating-base dynamics, payload changes,
external pushes, and terrain variations remain observable through their
effects on contact-force distribution and joint-space loading. Here,
contact-agnostic means that CARO captures the joint-space consequences of
contact interactions without explicitly estimating contact states,
locations, or wrenches. Rather than directly applying
the estimated residual as joint-level compensation, CARO provides it to
the policy as an internal-model adaptation signal.

CARO is trained under the same terrain, command, and domain-randomization conditions as the nominal policy, without specialized disturbance curricula or additional adaptation supervision. We evaluate the resulting policy through systematic payload-terrain sweeps and sudden payload changes in simulation. CARO consistently improves locomotion robustness relative to representative baselines based on standard RL, history-based adaptation, and adaptive control. Real-world experiments further demonstrate zero-shot traversal of unseen terrains, heading maintenance under center-of-mass shifts, increased payload capacity, and improved robustness during elevated-platform landings.

The main contributions of this work are summarized as follows:
\begin{itemize}
\item \textbf{Deployable contact-agnostic internal-model adaptation.}
We introduce a fixed-base Euler--Lagrange residual observer that can be directly integrated into a quadruped locomotion policy without explicit contact-force estimation, force/torque sensing, or measurements of the floating-base position and linear velocity.

\item \textbf{Policy-level use of structured residual feedback.}
Rather than using the observed residual for direct joint-level compensation, CARO provides it to the RL policy as an adaptation input, allowing the policy to determine an appropriate response under contact-rich and out-of-distribution conditions.

\item \textbf{Zero-shot robustness without specialized robustness training.}
Using the same training conditions as the nominal policy, CARO improves robustness to out-of-distribution payloads, center-of-mass shifts, terrain geometries, and abrupt dynamics changes in both simulation and real-world experiments.

\end{itemize}

\section{Related Work}

\subsection{Robustness in RL-Based Legged Locomotion}

Policy learning has become a dominant paradigm for legged locomotion, with large-scale simulation, curriculum learning, and domain randomization forming the foundation of robust policy training \cite{rudin_learning_2021}. To handle distribution shifts at deployment, a prominent line of work augments RL policies with latent adaptation modules or context encoders that infer hidden environment properties from proprioceptive histories \cite{kumar_rma_2021,nahrendra_dreamwaq_2023,xiao_safe_2024,long_hybrid_2024}. RMA \cite{kumar_rma_2021} trains a rapid adaptation module to reproduce a privileged environment representation online, while DreamWaQ \cite{nahrendra_dreamwaq_2023} extends history-based adaptation through imagination-augmented learning. Other methods learn control-relevant latent variables or history-dependent representations to improve robustness \cite{long_learning_2024,zhi_learning_2025}. Although effective, their adaptation signals are generally inferred implicitly from history rather than explicitly derived from a structured closed-loop residual.

A complementary direction incorporates control-inspired structures into learning-based policies
\cite{ji_concurrent_2022,lyu_rl2ac_2024,gao_neural_2025}. These methods introduce predictive errors, observer states, adaptive terms, or residual signals into the policy input. An estimator network \cite{ji_concurrent_2022} learns to predict unmeasured state information to improve policy performance, but the learned estimator is not guaranteed to provide structured feedback under out-of-distribution conditions.
RL2AC \cite{lyu_rl2ac_2024} combines learned context representations with adaptive control signals for robust legged locomotion but requires onboard torque sensing for its adaptive controller.
Neural-IMC \cite{gao_neural_2025} provides body-level model-prediction errors to the policy as auxiliary feedback. However, body-level prediction errors and generic adaptive signals do not directly provide the joint-space internal-model feedback needed to characterize contact-rich locomotion dynamics. Additionally, the requirement for floating-base state estimation limits the direct deployment of these methods on legged robots.
DeepDOB \cite{DeepDOB_locomotion} learns a disturbance observer to reject residual effects associated with predicted desired actions. Its results show improved locomotion-policy robustness by calibrating the current action with a desired action.

CARO constructs a lightweight joint-level residual observation using the fixed-base Euler--Lagrange internal model. The residual is provided to the policy as an adaptation signal, enabling structured feedback under out-of-distribution contact distribution and torque residuals without explicitly identifying contact forces.
ADAPT \cite{lyu2026ADAPT} similarly combines an Euler--Lagrange model and a momentum observer with an RL policy, using a two-phase training framework for force-aware humanoid locomotion. However, its objective is to provide the policy with accurate end-effector force information. It also requires a floating-base model and base-velocity estimates, which may not be available on all platforms. In contrast, CARO uses a fixed-base model, requires no root velocity estimation, and targets zero-shot robustness without specialized two-phase training.

\subsection{Disturbance Estimation for Legged Robots}

Contact-rich dynamics present unique challenges for disturbance estimation without contact-state or force/torque sensing.
Classical disturbance observers \cite{chenNonlinearDisturbanceObserver2000,chenDisturbanceObserverBased2004} and generalized-momentum observers \cite{de_luca_actuator_2003,robotcollision2017,de_luca_collision_2005} infer external joint torques from discrepancies between nominal robot dynamics and measured motion \cite{de_luca_collision_2005,haddadin_robot_2017}. Such methods have been applied to manipulators for contact estimation and physical interaction control \cite{magrini_control_2015,robotcollision2017}.

Recent work extends sensorless estimation to legged and humanoid robots. Proprioceptive external torque learning estimates external joint torques and contact wrenches for floating-base humanoids using internal measurements \cite{lim_proprioceptive_2023,concur_learn,zhi_learning_2025}, but generalization to unseen payloads and contact conditions remains challenging.
MOB-Net \cite{lim_mobnet_2025} combines a momentum observer with learned model-uncertainty calibration to improve external-torque estimation in high-dimensional humanoid robots.
Sensor-fusion approaches have also been developed to estimate effective joint torques on humanoid robots without dedicated torque sensors \cite{sorrentino_ukf_2024}.

Rather than recovering physically accurate external torques for force control, collision handling, or contact localization, CARO uses the observed joint-level residual as an adaptation input to an RL locomotion policy. This relaxes the need for precise disturbance identification and allows the policy to exploit internal-model feedback under contact-rich and out-of-distribution conditions.

\section{Contact-Agnostic Residual Observation}

\subsection{Preliminaries}

Consider a quadruped robot with a floating base and $n$ actuated joints.
Let $\bm{x}_b$ denote the floating-base configuration, and let
\begin{equation}
    \bm{\nu}
    =
    \begin{bmatrix}
        \bm{v}_b^\top &
        \dot{\bm{q}}^\top
    \end{bmatrix}^{\top}
    \in \mathbb{R}^{n+6},
\end{equation}
denote the generalized velocity, where
$\bm{v}_b \in \mathbb{R}^{6}$ is the floating-base spatial velocity and
$\bm{q},\dot{\bm{q}} \in \mathbb{R}^{n}$ are the actuated joint positions
and velocities. The full floating-base dynamics can be written as
\begin{equation}
    \bm{M}_{\mathrm{full}}(\bm{x}_b,\bm{q})\dot{\bm{\nu}}
    +
    \bm{h}_{\mathrm{full}}(\bm{x}_b,\bm{q},\bm{\nu})
    =
    \bm{S}^{\top}\bm{\tau}
    +
    \bm{J}_{c}^{\top}(\bm{x}_b,\bm{q})\bm{f}_{c}
    +
    \bm{\tau}_{\mathrm{ext}},
    \label{eq:floating_base_dynamics}
\end{equation}
where $\bm{M}_{\mathrm{full}}$ is the generalized inertia matrix,
$\bm{h}_{\mathrm{full}}$ is the nonlinear bias term containing the Coriolis
and gravitational terms, $\bm{\tau} \in \mathbb{R}^{n}$ is the actual
joint torque, $\bm{f}_{c}$ contains the stacked contact wrenches, and
$\bm{\tau}_{\mathrm{ext}}$ denotes additional generalized disturbances.
The actuation selection matrix is
\begin{equation}
    \bm{S}
    =
    \begin{bmatrix}
        \bm{0}_{n \times 6} &
        \bm{I}_{n}
    \end{bmatrix}.
\end{equation}

The generalized contact-force vector can be partitioned into its floating-base
and actuated-joint components:
\begin{equation}
    \bm{J}_{c}^{\top}\bm{f}_{c}
    =
    \begin{bmatrix}
        \bm{J}_{c,b}^{\top}\bm{f}_{c} \\
        \bm{J}_{c,j}^{\top}\bm{f}_{c}
    \end{bmatrix},
    \label{eq:contact_force_partition}
\end{equation}
where $\bm{J}_{c,b}^{\top}\bm{f}_{c}$ is the contact wrench projected
onto the floating base and
$\bm{J}_{c,j}^{\top}\bm{f}_{c}$ is the corresponding joint-space
loading.

The actuated rows of \eqref{eq:floating_base_dynamics} are
\begin{equation}
    \bm{M}_{jb}\dot{\bm{v}}_b
    +
    \bm{M}_{jj}\ddot{\bm{q}}
    +
    \bm{h}_{j}
    =
    \bm{\tau}
    +
    \bm{J}_{c,j}^{\top}\bm{f}_{c}
    +
    \bm{\tau}_{\mathrm{ext},j},
    \label{eq:actuated_floating_base_dynamics}
\end{equation}
where $\bm{M}_{jb}$ characterizes the dynamic coupling between the
floating base and actuated joints.

For example, a wrench $\bm{w}_b$ applied directly to the trunk can be
represented as
\begin{equation}
    \bm{\tau}_{\mathrm{ext}}
    =
    \begin{bmatrix}
        \bm{w}_b \\
        \bm{0}_{n}
    \end{bmatrix}.
    \label{eq:base_external_wrench}
\end{equation}
Although such a disturbance has no direct component in the actuated
coordinates, it affects the joint dynamics through the base-joint
coupling term $\bm{M}_{jb}\dot{\bm{v}}_b$ and through changes in the
contact wrenches $\bm{f}_{c}$.

The policy $\bm{\pi}_{\theta}$ outputs a reference joint-position increment
$\bm{q}_{r}$, which a low-level proportional-derivative (PD) controller
converts into a torque command:
\begin{equation}
    \bm{\tau}_{\text{cmd}}
    =
    \bm{K}_{p}(\bm{q}_{0}+\bm{q}_{r}-\bm{q})
    -
    \bm{K}_{d}\dot{\bm{q}},
    \label{eq:torque_control}
\end{equation}
where $\bm{q}_{0}$ is the default joint position and $\bm{q}_{r}$ is the
policy-generated reference joint-position increment.
$\bm{K}_{p}$ and $\bm{K}_{d}$ are positive-definite diagonal gain
matrices. 

\subsection{Fixed-Base Euler--Lagrange Internal Model}

Constructing an internal-model observer from the full dynamics in
\eqref{eq:floating_base_dynamics} requires an accurately calibrated
floating-base model and reliable estimates of the complete generalized
state, including the floating-base linear velocity. CARO instead
deliberately adopts a fixed-base Euler--Lagrange model as a lightweight
and deployable internal model. The base is locked only within the internal
model; the physical robot remains a floating-base system.
The resulting joint-space model is
\begin{equation}
    \bm{M}_{\mathrm{f}}(\bm{q})\ddot{\bm{q}}
    +
    \bm{h}_{\mathrm{f}}(\bm{q},\dot{\bm{q}})
    =
    \bm{\tau}_{\mathrm{cmd}}
    +
    \bm{r},
    \label{eq:fixed_base_dynamics}
\end{equation}
where $\bm{M}_{\mathrm{f}} \in \mathbb{R}^{n \times n}$,
$\bm{C}_{\mathrm{f}} \in \mathbb{R}^{n \times n}$, and
$\bm{G}_{\mathrm{f}} \in \mathbb{R}^{n}$ are, respectively, the inertia,
Coriolis, and gravity terms computed from the
fixed-base rigid-body model;
$\bm{h}_{\mathrm{f}} = \bm{C}_{\mathrm{f}}(\bm{q},\dot{\bm{q}})\dot{\bm{q}} + \bm{G}_{\mathrm{f}}(\bm{q}) \in \mathbb{R}^{n}$ is the nonlinear bias term; and $\bm{r} \in \mathbb{R}^{n}$ is the internal-model residual.
In \eqref{eq:residual_definition}, we use $\bm{\tau}_{\mathrm{cmd}}$ rather
than $\bm{\tau}$ to avoid the need for torque sensing during residual
estimation.
The internal-model residual is defined as
\begin{equation}
    \bm{r}
    :=
    \bm{M}_{\mathrm{f}}(\bm{q})\ddot{\bm{q}}
    +
    \bm{h}_{\mathrm{f}}(\bm{q},\dot{\bm{q}})
    -
    \bm{\tau}_{\mathrm{cmd}}.
    \label{eq:residual_definition}
\end{equation}

Comparing \eqref{eq:residual_definition} with the actuated floating-base
dynamics in \eqref{eq:actuated_floating_base_dynamics} gives
\begin{equation}
    \bm{r}
    =
    \bm{J}_{c,j}^{\top}\bm{f}_{c}
    +
    \bm{\tau}_{\mathrm{ext},j}
    +
    \bm{\Delta}_{\mathrm{fb}} + \bm{\delta}_{\tau},
    \label{eq:residual_interpretation}
\end{equation}
Here, the fixed-base discrepancy is
\begin{equation}
\begin{aligned}
    \bm{\Delta}_{\mathrm{fb}}
    :=
    {}&
    \left(
        \bm{M}_{\mathrm{f}}-\bm{M}_{jj}
    \right)\ddot{\bm{q}}
    -
    \bm{M}_{jb}\dot{\bm{v}}_b
    \\
    &+
    \bm{C}_{\mathrm{f}}(\bm{q},\dot{\bm{q}})\dot{\bm{q}}
    +
    \bm{G}_{\mathrm{f}}(\bm{q})
    -
    \bm{h}_{j}.
\end{aligned}
    \label{eq:fixed_base_discrepancy}
\end{equation}
The vector $\bm{\delta}_{\tau} := \bm{\tau} - \bm{\tau}_{\mathrm{cmd}}$
denotes the motor torque-tracking error.
CARO does not attempt to identify or separate the individual terms in
\eqref{eq:residual_interpretation}. Instead, their combined effect is
retained as a structured joint-space signal for policy adaptation.

% \begin{figure*}[t]
%     \centering
%     \includegraphics[
%         width=0.96\textwidth
%     ]{figures/fixed_base_residual_mechanism.pdf}
%     \caption{
%         Illustration of the fixed-base residual mechanism.
%         Left: in the full floating-base dynamics, a trunk payload or
%         external push affects locomotion through coupled base motion,
%         contact-wrench variation, and joint response.
%         Right: CARO locks the base only inside its internal model.
%         The resulting changes in contact-force distribution and omitted
%         floating-base coupling appear as a joint-space residual.
%         CARO does not reconstruct the individual contact forces or the
%         applied base wrench; it provides the observed residual directly
%         to the locomotion policy as an adaptation signal.
%     }
%     \label{fig:fixed_base_residual_mechanism}
% \end{figure*}

The effects of a base-level perturbation become observable in the fixed-base
residual. An attached payload changes the inertial and
gravitational loading of the floating base and alters the support forces
required from the stance legs. Similarly, an external push induces base
acceleration and a transient redistribution of the contact wrenches.
Although neither effect is explicitly represented by the fixed-base
model, both modify the joint-space response through
$\bm{J}_{c,j}^{\top}\bm{f}_{c}$ and
$\bm{\Delta}_{\mathrm{fb}}$, and therefore appear in the residual
$\bm{r}$.
Such residual signatures commonly arise when payload or terrain changes
redistribute loading among the front, rear, left, and right legs.

% This property is particularly relevant to multi-contact locomotion.
% Different contact-force distributions can produce the same resultant
% wrench on the floating base while inducing different joint-space
% loadings. 
% At a given robot configuration, it is possible to have
% \begin{equation}
% \begin{aligned}
%     \bm{J}_{c,b}^{\top}\bm{f}_{c}^{(1)}
%     &=
%     \bm{J}_{c,b}^{\top}\bm{f}_{c}^{(2)}, \\
%     \bm{J}_{c,j}^{\top}\bm{f}_{c}^{(1)}
%     &\neq
%     \bm{J}_{c,j}^{\top}\bm{f}_{c}^{(2)}.
% \end{aligned}
%     \label{eq:contact_redistribution_signature}
% \end{equation}
% Thus, two contact configurations may be indistinguishable from their net
% base wrench but remain distinguishable through their joint-space
% projections. 

CARO therefore does not explicitly recover contact states, locations, or
wrenches; instead, it retains their joint-space effects in the residual.

\subsection{Residual Observer}

Direct evaluation of \eqref{eq:residual_definition} requires joint
acceleration $\ddot{\bm{q}}$, which is generally unavailable and can be
noisy when obtained through numerical differentiation. We therefore
introduce a momentum-based disturbance-observer structure to estimate
$\bm{r}$ without explicitly measuring joint acceleration
\cite{chenDisturbanceObserverBased2004,de_luca_actuator_2003}.

Consider the first-order residual-estimation dynamics
\begin{equation}
    \dot{\hat{\bm{r}}}
    =
    \bm{L}
    \left(
        \bm{r}-\hat{\bm{r}}
    \right),
    \label{eq:ideal_residual_observer}
\end{equation}
where $\hat{\bm{r}} \in \mathbb{R}^{n}$ is the estimated residual and
$\bm{L} \in \mathbb{R}^{n \times n}$ is a positive-definite diagonal
observer-gain matrix, i.e., $\bm{L} \succ \bm{0}$.

Define the fixed-base momentum as
$\bm{p}_{\mathrm{f}} := \bm{M}_{\mathrm{f}}(\bm{q})\dot{\bm{q}}$.
Differentiating this expression gives
\begin{equation}
    \dot{\bm{p}}_{\mathrm{f}}
    =
    \bm{M}_{\mathrm{f}}(\bm{q})\ddot{\bm{q}}
    +
    \dot{\bm{M}}_{\mathrm{f}}(\bm{q})\dot{\bm{q}}.
    \label{eq:fixed_base_momentum_derivative}
\end{equation}

Using \eqref{eq:residual_definition} to replace
$\bm{M}_{\mathrm{f}}(\bm{q})\ddot{\bm{q}}$ in
\eqref{eq:fixed_base_momentum_derivative} yields
\begin{equation}
    \dot{\bm{p}}_{\mathrm{f}}
    =
    \bm{r} + \bm{\tau}_{\mathrm{cmd}} + \bm{b}_{\mathrm{f}},
    \label{eq:fixed_base_momentum_derivative_replaced}
\end{equation}
where $\bm{b}_{\mathrm{f}} := \dot{\bm{M}}_{\mathrm{f}}(\bm{q})\dot{\bm{q}} - \bm{h}_{\mathrm{f}}$ is the fixed-base momentum bias.
To eliminate the dependence on $\ddot{\bm{q}}$, we introduce the
auxiliary state
\begin{equation}
    \bm{z}
    =
    \hat{\bm{r}}
    -
    \bm{L}\bm{p}_{\mathrm{f}},\quad
    \dot{\bm{z}}
    = \dot{\hat{\bm{r}}} - \bm{L}\dot{\bm{p}}_{\mathrm{f}}.
    \label{eq:observer_auxiliary_state}
\end{equation}
Substituting
\eqref{eq:observer_auxiliary_state} and
\eqref{eq:fixed_base_momentum_derivative_replaced} into
\eqref{eq:ideal_residual_observer} yields the implementable residual observer
\begin{equation}
\begin{aligned}
    \dot{\bm{z}}
    &=
    -\bm{L}
    \big(
        \bm{z}
        +
        \bm{L}\bm{p}_{\mathrm{f}}
        +
        \bm{\tau}_{\mathrm{cmd}}
        +
        \bm{b}_{\mathrm{f}}
    \big), \\
    \hat{\bm{r}}_{\mathrm{eff}}
    &=
    \bm{z}
    +
    \bm{L}\bm{p}_{\mathrm{f}}.
\end{aligned}
    \label{eq:implementable_residual_observer}
\end{equation}
In addition to the fixed-base model, the observer requires only measured
joint positions and velocities and the torque command generated by
\eqref{eq:torque_control}. 
It therefore requires no force/torque sensors,
explicit contact estimation, or measurements of the floating-base position
and linear velocity. 
Let $\hat{\bm{b}}_{\mathrm{f}}$ denote the estimated momentum bias from any available method,
we approximate $\bm{b}_{\mathrm{f}}$ with $\hat{\bm{b}}_{\mathrm{f}}$ in
\eqref{eq:implementable_residual_observer}. 
The observer then outputs effective residual $\hat{\bm{r}}_{\mathrm{eff}}$ instead of $\hat{\bm{r}}$.
Details of the fixed-base model and residual-observer
implementation are provided in Appendix~\ref{app:implementation}.

\subsection{Boundedness of Residual Estimation}

The following result establishes a bound on the proposed residual observer's
estimation error.

\begin{theorem}
\label{thm:sampled_residual_bound}
Let $t_k=kT_s$ denote the sampling instants, and suppose that
\eqref{eq:implementable_residual_observer} is integrated using forward
Euler. For the analysis, define the Euler-consistent sampled residual
\begin{equation}
    \bm{\rho}_k
    :=
    \frac{\bm{p}_{\mathrm{f},k+1}-\bm{p}_{\mathrm{f},k}}{T_s}
    -\bm{\tau}_{\mathrm{cmd},k}
    -\hat{\bm{b}}_{\mathrm{f},k}.
    \label{eq:sampled_residual}
\end{equation}
Assume that $0<T_s\lambda_i(\bm{L})<2$ for every eigenvalue of
$\bm{L}$, and define
$\bm{A}:=\bm{I}-T_s\bm{L}$ and
$\alpha:=\lVert\bm{A}\rVert_2<1$.  If
\begin{equation}
    \lVert\bm{\rho}_k-\bm{\rho}_{k-1}\rVert_2\leq\beta,
    \qquad k\geq 1,
    \label{eq:sampled_residual_increment_bound}
\end{equation}
then the one-step-delayed estimation error
$\bm{e}_k:=\bm{\rho}_{k-1}-\hat{\bm{r}}_{\mathrm{eff},k}$ satisfies, for every
$k\geq 1$,
\begin{equation}
    \lVert\bm{e}_k\rVert_2
    \leq
    \alpha^{k-1}\lVert\bm{e}_1\rVert_2
    +\frac{\alpha(1-\alpha^{k-1})}{1-\alpha}\,\beta.
    \label{eq:sampled_residual_error_bound}
\end{equation}
Consequently,
\begin{equation}
    \limsup_{k\to\infty}\lVert\bm{e}_k\rVert_2
    \leq\frac{\alpha}{1-\alpha}\,\beta.
    \label{eq:sampled_residual_ultimate_bound}
\end{equation}
\end{theorem}
The proof is provided in Appendix~\ref{app:residual_bound_proof}.

\subsection{Integration with the RL Policy}
\label{sec:policy_integration}

Let $\bm{o}_{t}^{\mathrm{nom}}$ denote the observation used by the nominal
locomotion policy. CARO augments the nominal observation with the
estimated fixed-base residual:
\begin{equation}
    \bm{o}_{t}^{\mathrm{CARO}}
    =
    \begin{bmatrix}
        \left(\bm{o}_{t}^{\mathrm{nom}}\right)^{\top} &
        \hat{\bm{r}}_{\mathrm{eff},t}^{\top}
    \end{bmatrix}^{\top}.
    \label{eq:caro_policy_observation}
\end{equation}
The actor then generates the reference joint-position increment according to
\begin{equation}
    \bm{q}_{r}
    =
    \bm{\pi}_{\theta}
    \left(
        \bm{o}_{t}^{\mathrm{CARO}}
    \right).
    \label{eq:caro_policy}
\end{equation}

In our simulation implementation, the nominal observation contains joint
positions, joint velocities, the previous action, projected gravity,
velocity commands, base linear velocity, and base angular velocity.
% \begin{equation}
%     \bm{o}_{t}^{\mathrm{nom}}
%     =
%     \begin{bmatrix}
%         \bm{q}_{t}^{\top} &
%         \dot{\bm{q}}_{t}^{\top} &
%         \bm{a}_{t-1}^{\top} &
%         \bm{g}_{t}^{\top} &
%         \left(\bm{v}_{t}^{\mathrm{cmd}}\right)^{\top} &
%         \bm{v}_{t}^{\top} &
%         \bm{\omega}_{t}^{\top}
%     \end{bmatrix}^{\top}.
%     \label{eq:nominal_policy_observation}
% \end{equation}
The base-state quantities are not inputs to the fixed-base residual observer
in \eqref{eq:implementable_residual_observer}. The CARO module itself
requires only joint-side measurements and the torque command.
Note that base linear velocity is omitted from the real-world observations
because it is not directly measurable without an estimator. It is included
in the simulation observations to facilitate comparison with the baselines.

During training, the critic additionally receives privileged information:
\begin{equation}
    \bm{o}_{t}^{\mathrm{critic}}
    =
    \begin{bmatrix}
        \left(\bm{o}_{t}^{\mathrm{CARO}}\right)^{\top} &
        \left(\bm{o}_{t}^{\mathrm{priv}}\right)^{\top}
    \end{bmatrix}^{\top},
\end{equation}
where $\bm{o}_{t}^{\mathrm{priv}}$ includes contact states, friction
coefficients, payload mass, center-of-mass (CoM) displacement, motor strength
scales, and the randomized PD gains. These quantities are provided only to
the critic during training and are unavailable to the actor at deployment.

The residual observer is activated from the beginning of policy training,
allowing the policy to learn how the internal-model residual relates to
its closed-loop response. 

\section{Empirical Study}

We evaluate CARO through two simulation studies and four real-world
experiments.
First, we conduct systematic payload-terrain sweeps
to assess zero-shot robustness under simultaneous
changes in robot dynamics and contact conditions.
Second, we examine the transient response of each policy to an abrupt
payload change during locomotion.
Finally, we deploy CARO on a physical \emph{DeepRobotics Lite3}
quadruped robot and evaluate its performance on unseen terrains, under
center-of-mass shifts and increased payloads, and during elevated-platform
landings.

All methods are trained on \textit{flat terrain} with the same command
distribution, domain-randomization ranges, reward function, and set of
available privileged variables.
CARO does not use a broader randomization distribution, a specialized
disturbance curriculum, or additional adaptation supervision.
Its only augmentation over the nominal policy is the residual observation
introduced in Section~III\@.
The shared training environment and objective allow the comparison to
focus on the effect of different adaptation mechanisms on
out-of-distribution generalization.

The domain-randomization parameters used during training are summarized
in Table~\ref{tab:dr}, while the complete reward specification is provided
in Appendix~\ref{app:reward_terms} (Table~\ref{tab:reward}).
As shown in Table~\ref{tab:dr}, the training-time added-mass range is
limited to $[-1.0,\,3.0]\,\mathrm{kg}$, while the remaining randomized
quantities cover contact, center-of-mass, actuator, and low-level control
variations. As detailed in Table~\ref{tab:reward}, all policies are
optimized with the same velocity-tracking objectives and the same
stability, smoothness, and safety penalties. Thus, the payload and terrain
evaluations below test the generalization of the learned adaptation
mechanisms rather than differences in the training objectives or
randomization budgets.
Each method is trained for 8,000 iterations, requiring approximately
4 hours on a laptop equipped with a single NVIDIA GeForce RTX 4060 GPU\@.
Our training and evaluation environment is built on \emph{legged\_gym}
\cite{rudin2021leggedgym}, with \emph{Isaac Gym}
\cite{makoviychuk2021isaac} as the simulator.

\begin{table}[h!]
  \centering
  \caption{Domain randomization shared by all methods during training.}
  \renewcommand{\arraystretch}{1.2}
  \label{tab:dr}
  \begin{tabular}{l c c}
    \toprule[0.3mm]
    \textbf{Term} & \textbf{Range} & \textbf{Unit} \\
    \midrule
    Added base mass & $[-1.0,\,3.0]$ & kg \\
    Friction coefficient & $[0.1,\,1.25]$ & - \\
    CoM offset $x$ & $[-0.05,\,0.01]$ & m \\
    CoM offset $y$ & $[-0.03,\,0.03]$ & m \\
    CoM offset $z$ & $[-0.03,\,0.03]$ & m \\
    Motor strength scale & $[0.8,\,1.2]$ & - \\
    $K_p$ scale & $[0.8,\,1.2]$ & - \\
    $K_d$ scale & $[0.8,\,1.2]$ & - \\
    \bottomrule[0.3mm]
  \end{tabular}
\end{table}

\begin{figure*}[t!]
    \centering
    \includegraphics[
        width=0.96\textwidth
    ]{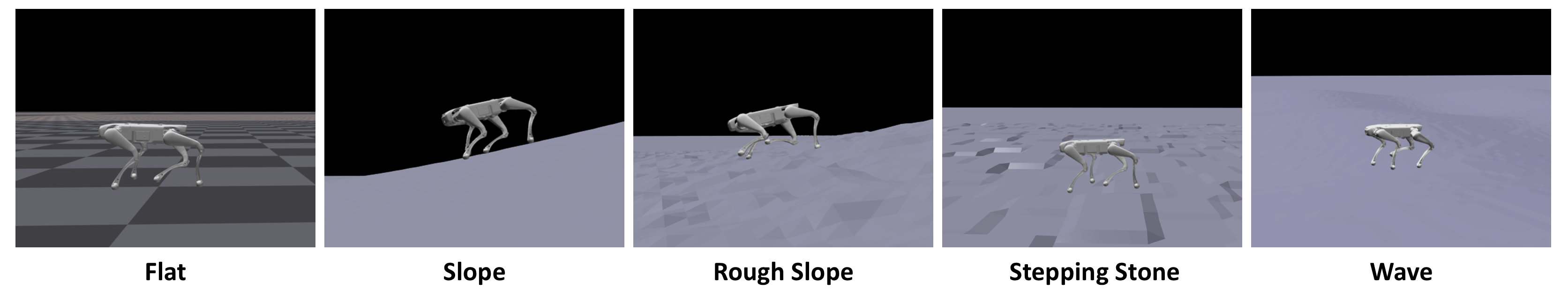}
    \caption{
        Simulation environments for zero-shot payload-terrain generalization.
        The robot is evaluated on five terrain types: \textit{flat}, \textit{slope}, \textit{rough slope}, \textit{stepping stones}, and \textit{wave}. All terrain difficulty levels are set to $0.7$.
    }
\end{figure*}

% RAL/IEEE two-column table; requires \usepackage{booktabs}.
\begin{table*}[h!]
  \centering
  \caption{Zero-shot payload-terrain robustness and tracking performance.}
  \label{tab:zeroshot_compact_tracking}
  \small
  \setlength{\tabcolsep}{2.9pt}
  \renewcommand{\arraystretch}{1.08}
  \begin{tabular}{@{}lccccccc@{\quad}cc@{\quad}cc@{\quad}cc@{}}
    \toprule
    & \multicolumn{6}{c}{Success by Terrain (\%) $\uparrow$}
    & \multicolumn{2}{c}{Success by Payload (\%) $\uparrow$}
    & \multicolumn{2}{c}{Aggregate Performance $\uparrow$}
    & \multicolumn{2}{c}{Tracking Error $\downarrow$} \\
    \cmidrule(lr){2-7}\cmidrule(lr){8-9}\cmidrule(lr){10-11}\cmidrule(l){12-13}
    Method
    & Flat & Slope & Rough & Stones & Wave & Mean
    & $2.5\times$ & $3.0\times$
    & \shortstack{Walk Dist.\\(m)} & \shortstack{Reward\\per Step}
    & \shortstack{Linear\\(m/s)} & \shortstack{Yaw Rate\\(rad/s)} \\
    \midrule
    Vanilla
    & 54.3 & 54.2 & 36.4 & 41.1 & 50.2 & 47.2
    & 1.0 & 0.0
    & 8.224 & 0.0128
    & 0.138 & 0.135 \\
    CARO
    & \textbf{94.9} & \textbf{94.4} & \underline{65.0}
    & \textbf{89.4} & \textbf{91.9} & \textbf{87.1}
    & \textbf{87.4} & \textbf{64.5}
    & \textbf{13.189} & \underline{0.0143}
    & \underline{0.105} & \textbf{0.080} \\
    RMA
    & \underline{92.6} & \underline{90.0} & \textbf{74.0}
    & \underline{76.5} & \underline{75.6} & \underline{81.7}
    & \underline{84.6} & \underline{36.4}
    & \underline{12.768} & \textbf{0.0145}
    & \textbf{0.097} & \underline{0.084} \\
    RL2AC
    & 70.3 & 72.7 & 34.7 & 69.4 & 59.0 & 61.2
    & 44.9 & 0.2
    & 9.752 & 0.0133
    & 0.112 & 0.122 \\
    \bottomrule
  \end{tabular}
  \vspace{-1mm}
  \begin{flushleft}
    \footnotesize \textbf{Boldface} denotes the best result;
    \underline{underlining} denotes the second-best result.
    ``Rough'' denotes rough slope and ``Stones'' denotes stepping stones. Each
    terrain column reports an average over the five mass scales in
    \eqref{eq:evaluation_mass_scales}; the two payload columns report selected
    high-payload slices averaged over the five terrains.
    Aggregate metrics are averaged over all 25 payload-terrain combinations.
  \end{flushleft}
\end{table*}

% Queue Fig. 4 early enough to place it on page 5.  The temporary counter
% change preserves the logical numbering while Fig. 3 remains later in the
% source so that it retains its original page-6 placement.
\setcounter{figure}{3}
\begin{figure}[t]
    \centering
    \includegraphics[width=0.95\linewidth]{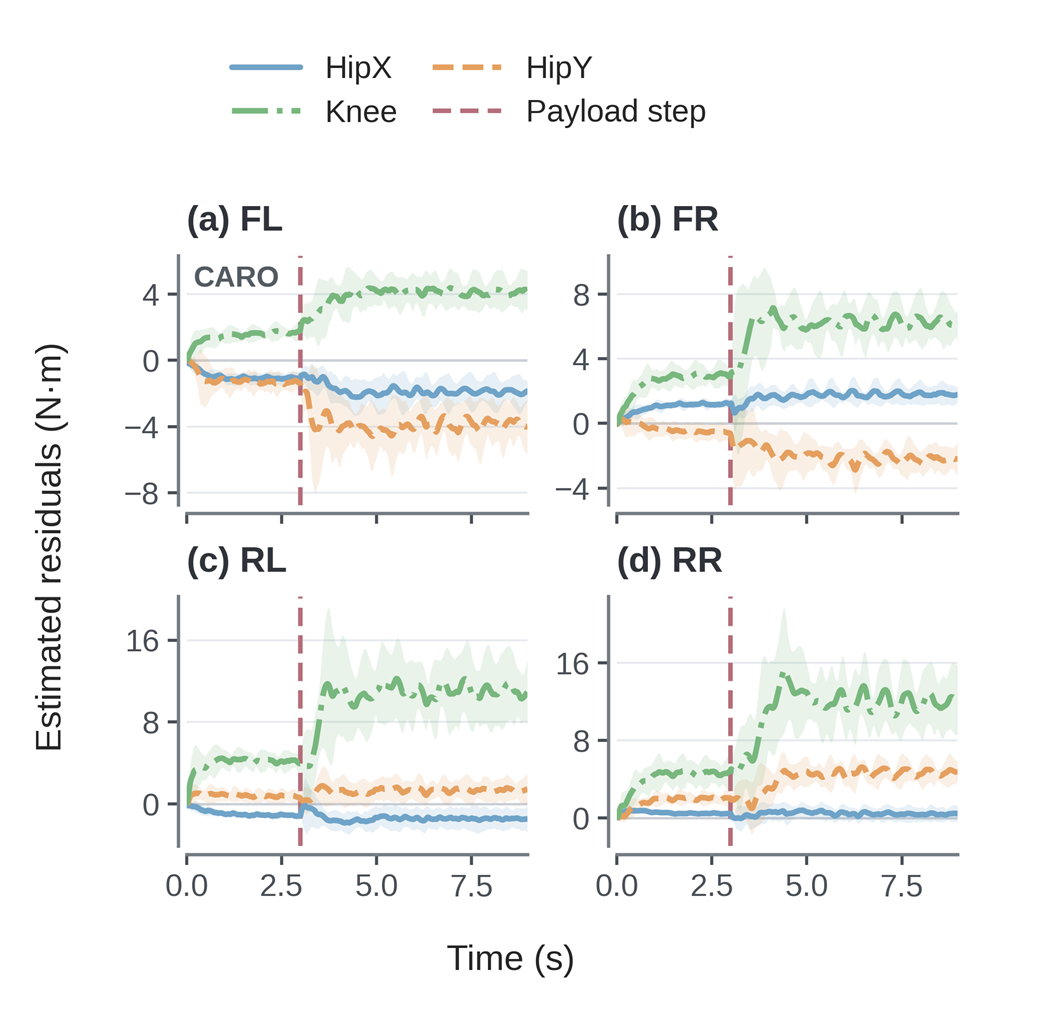}
    \caption{CARO residual estimates for the hip-$x$, hip-$y$, and knee
    joints of the (a) front-left (FL), (b) front-right (FR), (c) rear-left
    (RL), and (d) rear-right (RR) legs. Curves show trial means, shaded
    regions indicate trial-to-trial variability, and the vertical dashed
    line marks the payload step.}
    \label{fig:disturbance_estimates}
\end{figure}
\setcounter{figure}{2}

% Queue the two full-width figures just after page 4 is shipped.  This keeps
% Fig. 3 on page 6 and makes Fig. 5 the next full-width figure for page 7.
\afterpage{%
\begin{figure*}[t]
    \centering
    \includegraphics[width=0.9\linewidth]{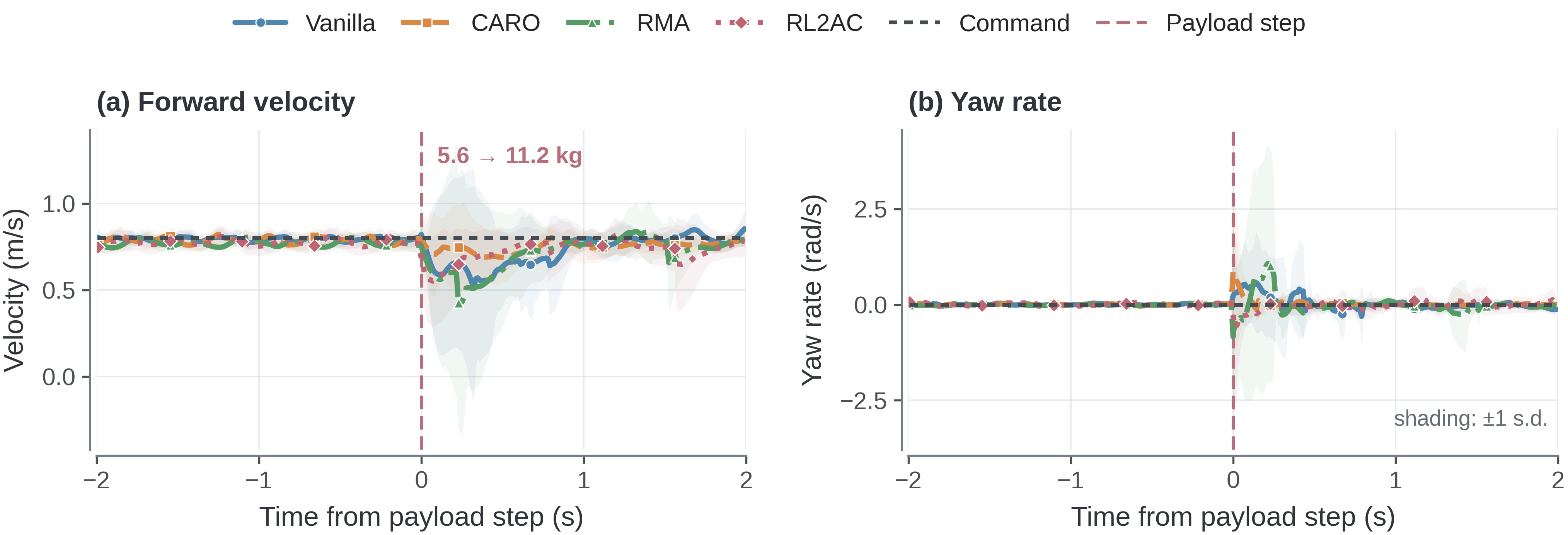}
    \caption{Response to the abrupt payload change: (a) forward velocity
    and (b) yaw rate. Curves show trial means, shaded regions denote one
    standard deviation, and the vertical dashed line at $t=3.0\,\mathrm{s}$
    marks the doubling of torso mass from $5.6$ to $11.2\,\mathrm{kg}$. The black
    dashed lines denote the commanded velocities.}
    \label{fig:velocity_response}
\end{figure*}

% Fig. 4 has already been numbered above; continue with Fig. 5 here.
\setcounter{figure}{4}
\begin{figure*}[t!]
    \centering
    \includegraphics[width=1.0\linewidth]{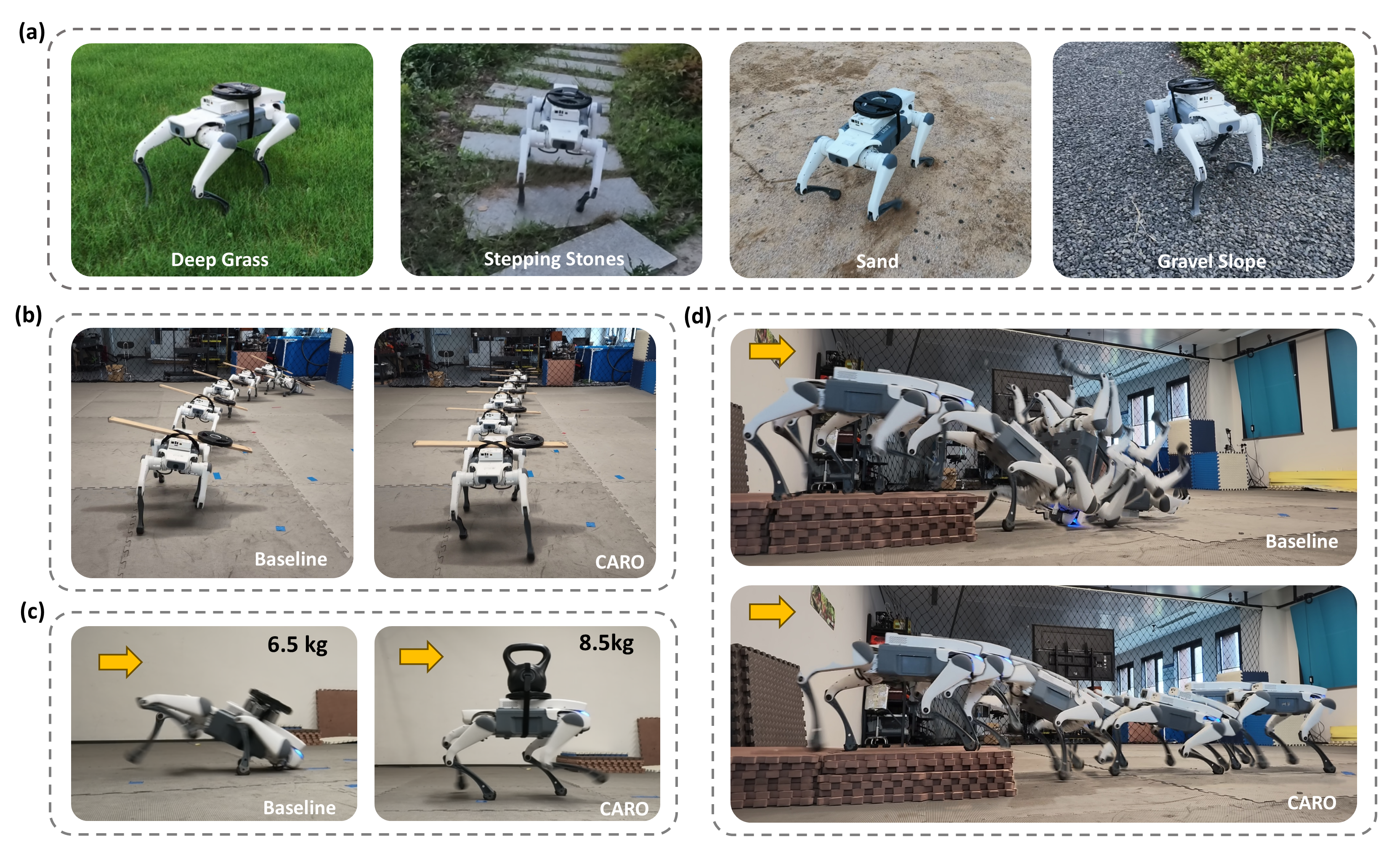}
    \caption{Real-world robustness experiments. (a) CARO traverses four
    representative unseen terrains while carrying a $2.5\,\mathrm{kg}$
    payload. (b) Time-composited trials with a
    $2.5\,\mathrm{kg}$ payload attached laterally: the Vanilla baseline deviates from the
    commanded heading, whereas CARO maintains it. (c) The Vanilla baseline
    falls with a $6.5\,\mathrm{kg}$ payload, whereas CARO maintains stable
    forward locomotion with an $8.5\,\mathrm{kg}$ payload. (d)
    Time-composited landings from a $0.2\,\mathrm{m}$ platform: the Vanilla
    baseline falls after impact, whereas CARO recovers and continues
    forward. Successive overlaid poses indicate temporal progression;
    yellow arrows in (c) and (d) denote the commanded travel direction.}
    \label{fig:real-world-exp}
\end{figure*}
}

\subsection{Zero-Shot Payload-Terrain Generalization}

We compare CARO against three representative baselines in
\emph{Isaac Gym}: a vanilla Proximal Policy Optimization (PPO) policy
(\emph{Vanilla}), the
teacher-student adaptation framework RMA
\cite{kumar_rma_2021}, and the adaptive-control-enhanced locomotion method RL2AC \cite{lyu_rl2ac_2024}.
The trained policies are evaluated without fine-tuning under
out-of-distribution payload and terrain conditions. We refer
to Appendix~\ref{app:baseline_implementation} for details of the baseline
implementations.

For each payload-terrain combination, we evaluate a fixed checkpoint of each
trained policy across 1,024 independently initialized parallel environments.
We define the success rate as the fraction of 15-second rollouts in which the
robot traverses $10\,\mathrm{m}$ without falling.

We evaluate each policy on five terrain types:
\emph{flat}, \emph{slope}, \emph{rough slope},
\emph{stepping stones}, and \emph{wave}.
Only the \emph{flat} terrain is used for training; the remaining terrain
geometries are unseen.
For each terrain type, the base-mass multiplier is selected from
\begin{equation}
    \rho_m
    \in
    \left\{
        1.0,\,
        1.5,\,
        2.0,\,
        2.5,\,
        3.0
    \right\},
    \label{eq:evaluation_mass_scales}
\end{equation}
where $\rho_m=1.0$ corresponds to the nominal base mass
($5.6\,\mathrm{kg}$).
For the mass sweep in \eqref{eq:evaluation_mass_scales}, the flat-terrain
evaluation assesses generalization to payload changes in isolation, whereas
the unseen-terrain evaluations jointly assess robustness to variations in
dynamics and terrain-induced changes in contact conditions.

Table~\ref{tab:zeroshot_compact_tracking} summarizes the sweep using
terrain-wise marginal success rates, two representative high-payload
slices, and metrics aggregated over all 25 payload-terrain combinations.
CARO achieves the highest mean success rate, $88.6\%$.
The advantage is most evident in the largest tabulated payload slice.
At $3.0\times$ nominal base mass, CARO retains a $64.5\%$ success rate,
compared with $36.4\%$ for RMA, $0.2\%$ for RL2AC, and $0\%$ for Vanilla.
At $2.5\times$ mass, CARO exceeds RMA by $2.8$ percentage points
($87.4\%$ versus $84.6\%$); at $3.0\times$, this margin widens to $28.1$
percentage points ($64.5\%$ versus $36.4\%$). It ranks first on four of the
five terrains; the exception is the rough slope, on which RMA obtains
$74.0\%$ and CARO obtains $65.0\%$. RMA's teacher-student framework may
provide a more direct adaptation signal for this particular terrain.

The aggregate metrics in Table~\ref{tab:zeroshot_compact_tracking} show
the robustness-tracking trade-off from a complementary perspective.
CARO travels the longest mean distance ($13.189\,\mathrm{m}$) and has the
lowest yaw-rate error ($0.080\,\mathrm{rad/s}$). RMA achieves a slightly
higher reward per step ($0.0145$ versus $0.0143$) and a slightly lower
linear-velocity error ($0.097\,\mathrm{m/s}$ versus $0.105\,\mathrm{m/s}$),
whereas CARO maintains a substantially higher success rate at the heaviest
tabulated payload. These results therefore do not imply uniform dominance
on every metric; rather, they indicate that the residual observation
particularly improves task-completion robustness under large, simultaneous
shifts in dynamics and contact conditions.

\subsection{Reaction to Sudden Payload Changes}

We next evaluate how rapidly each policy responds to an abrupt change in
robot dynamics. We initialize 100 environments with the robot walking
forward on flat terrain at a commanded velocity of $0.8\,\mathrm{m/s}$.
At $t=3.0\,\mathrm{s}$, the torso mass of each robot is instantaneously
doubled from $5.6\,\mathrm{kg}$ to $11.2\,\mathrm{kg}$ by adding a payload to
the robot's back while locomotion continues. The added $5.6\,\mathrm{kg}$
equals $50\%$ of the robot's nominal total mass and exceeds the maximum
$3.0\,\mathrm{kg}$ added mass used during training (Table~\ref{tab:dr}).

Figure~\ref{fig:velocity_response} compares the transient command-tracking
responses of all four policies. Before the step, their mean forward
velocities all converge to the $0.8\,\mathrm{m/s}$ command. Immediately
after the step, Vanilla and especially RMA exhibit a larger forward-speed
drop and a wider trial-to-trial spread in
Figure~\ref{fig:velocity_response}(a), whereas CARO's mean forward velocity
remains closer to the command and recovers quickly, with a comparatively
narrow variability band. All methods
show a transient yaw-rate response in Figure~\ref{fig:velocity_response}(b),
but CARO's mean yaw rate quickly returns to the zero command. RMA has the largest
initial dispersion, while Vanilla and RL2AC exhibit larger fluctuations later
in the trial.

The corresponding residual estimates are shown in
Figure~\ref{fig:disturbance_estimates}. The observed change is not a uniform
offset shared by all twelve joints. In particular, the knee residual
increases on every leg, with the largest sustained changes on the rear legs
in panels (c) and (d) of Figure~\ref{fig:disturbance_estimates}.
% Meanwhile,
% the hip-$y$ residual becomes more negative on the front legs but more
% positive on the rear legs, and the hip-$x$ channels show smaller
% leg-dependent shifts.
The increased post-step variability, most visibly in the rear-knee
channels, also reflects trial-dependent transient loading. These spatially
distinct signatures are consistent with a redistribution of support loads
and give the policy more information than a scalar payload flag, without
requiring CARO to identify either the payload mass or individual contact
forces. The nonzero values before the step are expected because the
residual also contains nominal contact loading and the fixed-base discrepancy
defined in \eqref{eq:residual_interpretation}.

Figure~\ref{fig:tracking_error_payload} aggregates the transient traces.
Among the plotted post-step means, CARO has the lowest error in both
panels. CARO's linear-velocity error increases only slightly, from roughly
$0.08\,\mathrm{m/s}$ to $0.09\,\mathrm{m/s}$, whereas Vanilla's error rises
from about $0.075\,\mathrm{m/s}$ to $0.125\,\mathrm{m/s}$ and RL2AC's rises
from about $0.09\,\mathrm{m/s}$ to $0.14\,\mathrm{m/s}$; RMA's remains near
$0.09\,\mathrm{m/s}$ but exhibits greater uncertainty. The yaw-rate
comparison is more pronounced: CARO's error increases from approximately
$0.05\,\mathrm{rad/s}$ to $0.07\,\mathrm{rad/s}$, whereas the post-step means
for Vanilla, RMA, and RL2AC are about $0.18\,\mathrm{rad/s}$,
$0.13\,\mathrm{rad/s}$, and $0.22\,\mathrm{rad/s}$, respectively. These
results show that CARO can detect and respond to an abrupt shift in dynamics
without an explicit payload estimator or online parameter identification.

\subsection{Real-World Experiments}

We deploy CARO on a physical \emph{DeepRobotics Lite3} quadruped
without hardware-specific policy fine-tuning.
To provide temporal context, we supply the policy with a history of the five most recent observations, 
rather than a single observation. We found this modification to be critical for reducing
vibration on the real robot. Additional details on the sim-to-real observer setup are
provided in Appendix~\ref{app:implementation}.
The experiments evaluate four complementary aspects of deployment
robustness: generalization across contact surfaces, adaptation to
asymmetric mass distributions, maximum payload capacity, and recovery
from high-impact landings.
Figure~\ref{fig:real-world-exp} summarizes representative trials from
these four real-world evaluations.

\subsubsection{Zero-Shot Multi-Terrain Traversal}

The robot is commanded to traverse flat ground, a gravel-covered slope,
sand, grass, and stepping stones while carrying a
$2.5\,\mathrm{kg}$ payload.
Of these terrain conditions, only flat ground is included during policy
training.
Representative trials on the four unseen terrains are shown in
Figure~\ref{fig:real-world-exp}(a).
CARO maintains stable locomotion across all tested surfaces without
terrain-specific adaptation or fine-tuning.
The results demonstrate that the estimated fixed-base residual provides useful
feedback under substantial changes in both payload and foot-ground
interaction.

\subsubsection{Center-of-Mass Shifts}

To evaluate robustness to asymmetric mass distributions, we attach a
$2.5\,\mathrm{kg}$ payload to one side of the robot and command it to
walk forward at $1\,\mathrm{m/s}$ on flat ground.
We repeat the evaluation three times for each policy.
This configuration introduces both a payload increase and a substantial
lateral center-of-mass shift.
The time-composited trials in Figure~\ref{fig:real-world-exp}(b) visualize
the resulting locomotion trajectories.
CARO rapidly restores a near-level body attitude and maintains the commanded
forward heading.
In contrast, the Vanilla policy develops a persistent roll toward the
loaded side and exhibits substantial heading deviation.
This experiment indicates that the residual observation allows the policy
to react not only to changes in total mass but also to asymmetries in load
distribution.

\subsubsection{Maximum Payload Capacity}

We incrementally increase the payload by 0.5 kg attached to the base until
the policy can no longer complete the prescribed locomotion trial without
falling.
Under this test protocol, CARO maintains stable forward locomotion with
an $8.5\,\mathrm{kg}$ payload, as shown in
Figure~\ref{fig:real-world-exp}(c).
The Vanilla policy, in comparison, fails to complete the trial with a
$6.5\,\mathrm{kg}$ payload.
The increased payload capacity provides direct hardware evidence that
CARO extends the operating range of the locomotion policy beyond the
payload distribution used during training.

\subsubsection{Elevated-Platform Landings}

Finally, we evaluate recovery from high-impact contact transitions using
repeated elevated-platform landing trials.
The robot is released from a $0.2\,\mathrm{m}$-high platform and
lands on flat ground using a policy trained only for flat-terrain
locomotion. We repeat the test three times for each policy, CARO achieves 3/3
successful landings, whereas the Vanilla policy falls in all three trials.
Figure~\ref{fig:real-world-exp}(d) compares the corresponding
time-composited landing sequences.
CARO absorbs the landing impact and rapidly regains a stable body
attitude, whereas the Vanilla policy frequently falls or exhibits
prolonged post-impact instability.
This experiment demonstrates that the residual observation improves
robustness to abrupt contact impulses and large transient deviations,
despite the absence of elevated-platform landing scenarios during
training.

\section{Conclusion}

This paper presented CARO, a contact-agnostic residual observation framework
that embeds a fixed-base Euler--Lagrange internal model in an RL locomotion
policy. The observer converts the mismatch between the modeled joint dynamics
and commanded joint torques into a structured adaptation input rather than
using it for direct torque compensation. It thereby retains the joint-space
consequences of changes in payload and contact conditions without explicitly
estimating contact states or wrenches. Under the sampling and
bounded-increment conditions of
Theorem~\ref{thm:sampled_residual_bound}, the resulting residual-estimation
error is ultimately bounded.

Under matched training conditions, CARO achieved an $88.6\%$ mean success
rate across the 25 payload-terrain combinations and retained a
$64.5\%$ success rate at $3.0\times$ nominal base mass. It also produced the
lowest plotted post-change tracking errors following a sudden
$5.6\,\mathrm{kg}$ payload addition, equal to $50\%$ of the nominal total
mass. On the physical \emph{DeepRobotics Lite3}, CARO
traversed four unseen terrains with a $2.5\,\mathrm{kg}$ payload, maintained
the commanded heading under a lateral center-of-mass shift, and sustained
forward locomotion with an $8.5\,\mathrm{kg}$ payload. By comparison, the
Vanilla policy failed to complete its trial with a $6.5\,\mathrm{kg}$ payload.
CARO also recovered after landing from a $0.2\,\mathrm{m}$-high platform.
Together, these results show
that a deliberately simplified internal model can provide useful
deployment-time feedback for out-of-distribution changes in dynamics and
contact conditions without online payload or contact identification.

\section{Limitations and Open Problems} \label{sec:discussion}

Several limitations remain. \textbf{First}, the boundedness result applies to
the residual-estimation error when the Euler-consistent sampled residual has
bounded step-to-step increments. It does not establish the stability of the
complete learned closed-loop system.
The observer gain also introduces a trade-off between response bandwidth and
sensitivity to measurement noise. A higher gain allows the observer to track
faster residual changes, but it also amplifies noise in the joint-position and
joint-velocity measurements.
\textbf{Second}, the residual is intentionally lumped and does not provide
physically identifiable contact forces, disturbance locations, or model
parameters. During highly dynamic motions, large floating-base accelerations
may also cause the fixed-base approximation error to dominate the
observation. Future work will investigate adaptive residual filtering and
evaluate the method on highly dynamic motions and under a broader range of
actuator and contact perturbations.

% Keep Fig. 6 at its original page-9 position after moving Fig. 5 forward.
\afterpage{%
\begin{figure}[t]
    \centering
    \includegraphics[width=0.9\linewidth]{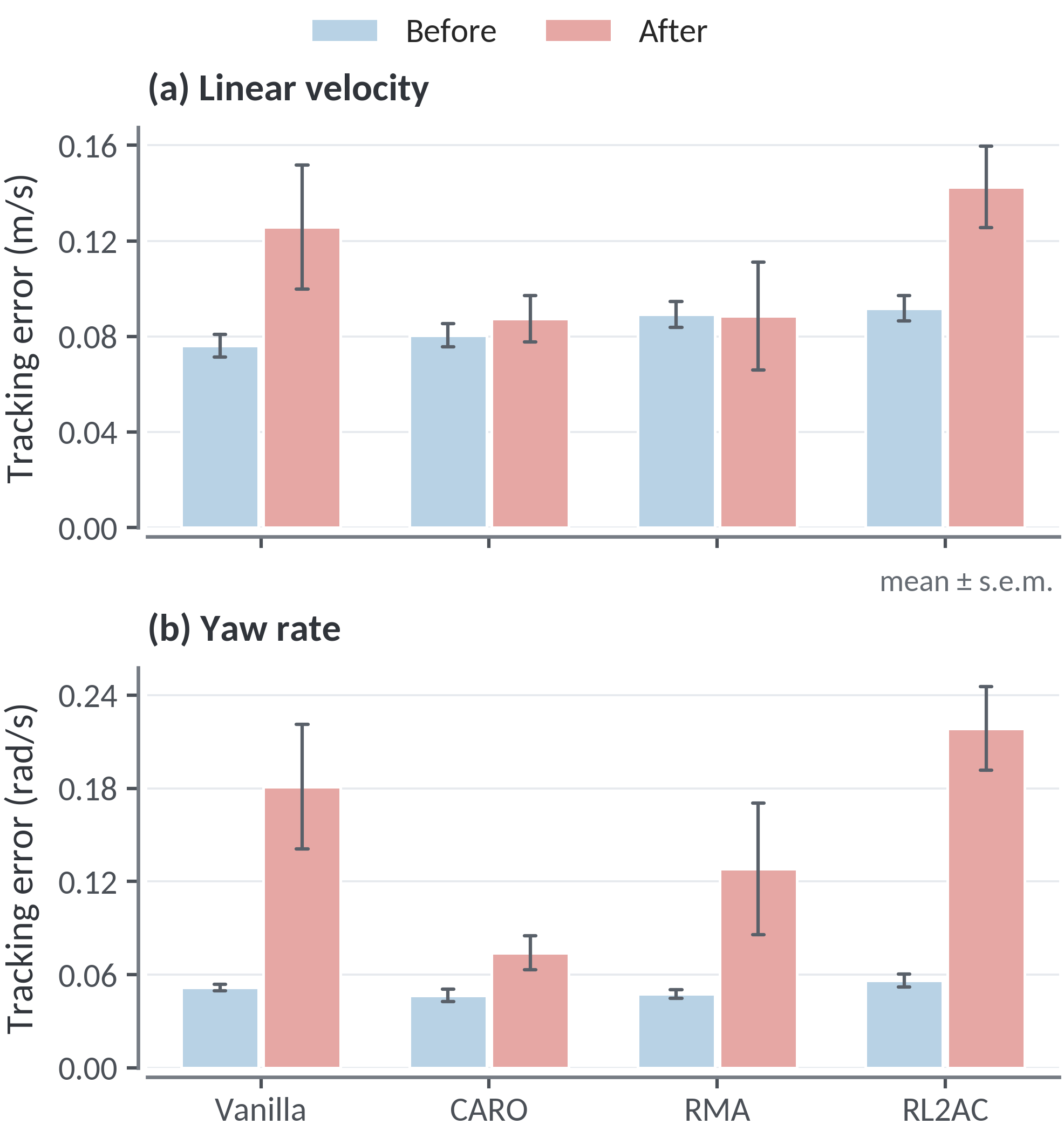}
    \caption{Mean (a) linear-velocity and (b) yaw-rate tracking errors over
    the pre-step and post-step intervals. Error bars denote the standard
    error of the mean.}
    \label{fig:tracking_error_payload}
\end{figure}
}

%%%%%%%%%%%%%%%%%%%%%%%%%%%%%%%%%%%%%%%%%%%%%%%%%%%%%%%%%%%%%%%%%%%%%%%%%%%%%%%%
\appendices

\section{Proof of the Residual-Estimation Bound}
\label{app:residual_bound_proof}

Applying forward Euler to the auxiliary-state dynamics in
\eqref{eq:implementable_residual_observer} gives
\begin{equation}
\begin{aligned}
    \bm{z}_{k+1}
    ={}&
    \bm{z}_k
    -T_s\bm{L}
    \big(
        \bm{z}_k
        +\bm{L}\bm{p}_{\mathrm{f},k}
        +\bm{\tau}_{\mathrm{cmd},k}
        +\hat{\bm{b}}_{\mathrm{f},k}
    \big).
    \label{eq:appendix_euler_auxiliary_update}
\end{aligned}
\end{equation}
Using
$\hat{\bm{r}}_{\mathrm{eff},k}=\bm{z}_k+\bm{L}\bm{p}_{\mathrm{f},k}$ and the
definition of $\bm{\rho}_k$ in \eqref{eq:sampled_residual}, we obtain
\begin{equation}
\begin{aligned}
    \hat{\bm{r}}_{\mathrm{eff},k+1}
    ={}&
    \bm{z}_{k+1}+\bm{L}\bm{p}_{\mathrm{f},k+1} \\
    ={}&
    \hat{\bm{r}}_{\mathrm{eff},k}
    +\bm{L}
        (\bm{p}_{\mathrm{f},k+1}-\bm{p}_{\mathrm{f},k}) \\
    &
    -T_s\bm{L}
        (\hat{\bm{r}}_{\mathrm{eff},k}
        +\bm{\tau}_{\mathrm{cmd},k}
        +\hat{\bm{b}}_{\mathrm{f},k}) \\
    ={}&
    \bm{A}\hat{\bm{r}}_{\mathrm{eff},k}
    +(\bm{I}-\bm{A})\bm{\rho}_k.
    \label{eq:appendix_discrete_residual_filter}
\end{aligned}
\end{equation}
Thus, \eqref{eq:appendix_discrete_residual_filter} is an exact algebraic
description of the implemented Euler update. The use of
$\bm{p}_{\mathrm{f},k+1}$ in \eqref{eq:sampled_residual} is limited to this
analysis; the online update \eqref{eq:appendix_euler_auxiliary_update}
remains causal and does not require a momentum difference.

From the definition
$\bm{e}_k=\bm{\rho}_{k-1}-\hat{\bm{r}}_{\mathrm{eff},k}$ and
\eqref{eq:appendix_discrete_residual_filter},
\begin{equation}
\begin{aligned}
    \bm{e}_{k+1}
    ={}&
    \bm{\rho}_k-\hat{\bm{r}}_{\mathrm{eff},k+1} \\
    ={}&
    \bm{A}(\bm{\rho}_k-\hat{\bm{r}}_{\mathrm{eff},k}) \\
    ={}&
    \bm{A}
    \big[
        \bm{e}_k+(\bm{\rho}_k-\bm{\rho}_{k-1})
    \big].
    \label{eq:appendix_error_recursion}
\end{aligned}
\end{equation}
Iterating \eqref{eq:appendix_error_recursion} yields
\begin{equation}
    \bm{e}_k
    =
    \bm{A}^{k-1}\bm{e}_1
    +
    \sum_{j=1}^{k-1}
    \bm{A}^{k-j}(\bm{\rho}_j-\bm{\rho}_{j-1}).
    \label{eq:appendix_error_solution}
\end{equation}
Because $\bm{L}$ is positive definite and diagonal, $\bm{A}$ is
symmetric, and the condition
$0<T_s\lambda_i(\bm{L})<2$ implies
\begin{equation}
    \alpha
    =\lVert\bm{A}\rVert_2
    =\max_i\lvert1-T_s\lambda_i(\bm{L})\rvert
    <1.
    \label{eq:appendix_contraction_factor}
\end{equation}
Taking the Euclidean norm of \eqref{eq:appendix_error_solution} and using
submultiplicativity together with the bound in
\eqref{eq:sampled_residual_increment_bound}, we obtain
\begin{equation}
\begin{aligned}
    \lVert\bm{e}_k\rVert_2
    \leq{}&
    \alpha^{k-1}\lVert\bm{e}_1\rVert_2
    +\beta\sum_{j=1}^{k-1}\alpha^{k-j} \\
    ={}&
    \alpha^{k-1}\lVert\bm{e}_1\rVert_2
    +\frac{\alpha(1-\alpha^{k-1})}{1-\alpha}\,\beta,
\end{aligned}
\end{equation}
which proves \eqref{eq:sampled_residual_error_bound}. Finally,
$\alpha^{k-1}\to0$ as $k\to\infty$; taking the limit superior then proves
\eqref{eq:sampled_residual_ultimate_bound}.

\section{Reward Terms}
\label{app:reward_terms}

All policies use the reward terms and weights listed in
Table~\ref{tab:reward}.

\begin{table}[h!]
  \centering
  \caption{Reward terms used for policy training.}
  \renewcommand{\arraystretch}{1.2}
  \label{tab:reward}
  \scalebox{0.94}{
    \begin{tabular}{l c c}
      \toprule[0.3mm]
      \textbf{Name} & \textbf{Expression} & \textbf{Weight} \\
      \midrule
      Linear-velocity tracking
      &
      $\exp\left(
      -\|\bm{v}_{xy}^{\mathrm{cmd}}-\bm{v}_{xy}\|^2/\sigma
      \right)$
      &
      $1.0$
      \\
      Yaw-rate tracking
      &
      $\exp\left(
      -|\omega_z^{\mathrm{cmd}}-\omega_z|^2/\sigma
      \right)$
      &
      $0.5$
      \\
      Vertical velocity
      &
      $v_z^2$
      &
      $-2.0$
      \\
      Roll/pitch angular velocity
      &
      $\|\bm{\omega}_{xy}\|^2$
      &
      $-0.05$
      \\
      Base orientation
      &
      $\|\bm{g}_{xy}\|^2$
      &
      $-0.2$
      \\
      Base height
      &
      $(h-0.36)^2$
      &
      $-1.0$
      \\
      Torque penalty
      &
      $\|\bm{\tau}\|^2$
      &
      $-1.0\times10^{-5}$
      \\
      Joint acceleration
      &
      $\|(\dot{\bm{q}}_{t}-\dot{\bm{q}}_{t-1})/\Delta t\|^2$
      &
      $-2.5\times10^{-7}$
      \\
      Action rate
      &
      $\|\bm{a}_{t}-\bm{a}_{t-1}\|^2$
      &
      $-0.01$
      \\
      Collision
      &
      $\sum_i
      \mathbb{I}
      \left(
      \|\bm{f}_{i}^{\mathrm{contact}}\|>0.1
      \right)$
      &
      $-1.0$
      \\
      Joint position limit
      &
      $\sum_i
      \mathrm{clip}
      \left(
      q_i-q_i^{\mathrm{lim}}
      \right)$
      &
      $-10.0$
      \\
      Foot air time
      &
      $\sum_i
      (t_{i}^{\mathrm{air}}-0.5)
      \mathbb{I}_{i}^{\mathrm{first}}$
      &
      $1.0$
      \\
      Standstill
      &
      $\|\bm{q}_{\mathrm{pen}}
      -\bm{q}_{\mathrm{pen}}^{\mathrm{default}}\|_1$
      &
      $-0.05$
      \\
      Foot slip velocity
      &
      $\sum_i
      \mathbb{I}_{i}^{\mathrm{contact}}
      \|\bm{v}_{i}^{\mathrm{foot}}\|$
      &
      $-0.05$
      \\
      \bottomrule[0.3mm]
    \end{tabular}
  }
\end{table}

\section{Implementation Details}
\label{app:baseline_implementation}
\subsection{Baseline Implementations}
For all methods, both the actor and critic use multilayer perceptrons with
hidden-layer dimensions $[512,\,256,\,128]$. Each deployable actor receives
its method-specific observation, whereas the critic additionally receives
privileged information, as described in
Section~\ref{sec:policy_integration}. Domain randomization
and reward terms are shared across all methods, as described in
Tables~\ref{tab:dr} and~\ref{tab:reward}. The baselines are configured as
follows.
\begin{itemize}
    \item The Vanilla policy is a plain multilayer perceptron (MLP) without
    an adaptation mechanism.
    \item The RMA baseline is implemented following the original paper
    \cite{kumar_rma_2021}. The teacher encoder is a two-layer MLP with
    hidden-layer dimensions $[256,\,128]$ and a latent dimension of $32$.
    The student encoder has the same architecture as the teacher encoder and
    is trained to predict the latent representation from a 25-step
    observation history using a mean-squared-error loss. We deploy the student encoder as the adaptation module in the actor.
    \item The RL2AC baseline is implemented following the original paper
    \cite{lyu_rl2ac_2024}. Its context-aided estimator network (CENet) takes
    a temporal history of proprioceptive observations and uses a two-layer
    encoder with hidden-layer dimensions $[128,\,64]$. Separate linear heads
    parameterize the means and log variances of the estimated body velocity,
    the next-step joint position $\bm{q}_{r}$, and the latent context. The resulting
    code is passed to a two-layer decoder with hidden-layer dimensions
    $[64,\,128]$ to reconstruct a single observation step. CENet is trained
    jointly with PPO using mean-squared-error losses for the velocity and
    next-step joint-position estimates, together with the reconstruction and
    Kullback--Leibler losses of a $\beta$-variational autoencoder. During
    control, the actor receives the estimated velocity and latent context and
    outputs the action $\bm{a}$, while $\bm{q}_{r}$ serves as the tracking
    reference for the adaptive controller. Following the RL2AC formulation,
    the compensating torque
    $\widehat{\bm{K}}_{u}(\bm{a}-\bm{q}_{r})$ is added to the nominal
    PD torque, and $\widehat{\bm{K}}_{u}$ is updated online using the original
    bounded-gain-forgetting composite adaptive law.
\end{itemize}

\subsection{Realization of the Residual Observer}
\label{app:implementation}
The fixed-base residual observer requires the computation of fixed-base
dynamics terms and robot-model Jacobians in
\emph{legged\_gym}. We compute $\bm{M}_{\mathrm{f}}$ and
$\bm{G}_{\mathrm{f}}$ in
PyTorch using the Jacobians provided by the \emph{Isaac Gym} API\@.
However, the current version of \emph{Isaac Gym} does not support the
computation of $\bm{C}_{\mathrm{f}}$. We therefore approximate
$\bm{C}_{\mathrm{f}}$ using the
diagonal relation
$\operatorname{diag}(\bm{C}_{\mathrm{f}}) =
0.1 \cdot \operatorname{diag}(\bm{M}_{\mathrm{f}})$.
This approximation is sufficient for the residual observer to provide useful
feedback to the policy and to support batched training in simulation. We
integrate the residual observer using the forward Euler method with the same
time step as the simulation, $T_s=\Delta t=0.02\,\mathrm{s}$. Higher-order
integration methods can improve residual-estimation accuracy,
but they are not necessary for the current implementation.
All diagonal entries of the observer-gain matrix $\bm{L}$ are set to $2.0$.

For real-world deployment, we use the \emph{Pinocchio} library
\cite{carpentier2019pinocchio} to compute the Jacobians of the observer model.
We use \emph{rl\_sar} \cite{fan-ziqi2024rl_sar} to implement the residual
observer alongside a LibTorch-compiled policy network for real-time
inference at $50\,\mathrm{Hz}$.

% \section*{ACKNOWLEDGMENT}

%%%%%%%%%%%%%%%%%%%%%%%%%%%%%%%%%%%%%%%%%%%%%%%%%%%%%%%%%%%%%%%%%%%%%%%%%%%%%%%%

\bibliographystyle{IEEEtran}
\bibliography{reference}  % .bib

@misc{long_hybrid_2024,
	title = {Hybrid {Internal} {Model}: {Learning} {Agile} {Legged} {Locomotion} with {Simulated} {Robot} {Response}},
	shorttitle = {Hybrid {Internal} {Model}},
	doi = {10.48550/arXiv.2312.11460},
	urldate = {2026-02-06},
	publisher = {arXiv},
	author = {Long, Junfeng and Wang, Zirui and Li, Quanyi and Gao, Jiawei and Cao, Liu and Pang, Jiangmiao},
	month = jan,
	year = {2024},
}

@article{ji_concurrent_2022,
	title = {Concurrent {Training} of a {Control} {Policy} and a {State} {Estimator} for {Dynamic} and {Robust} {Legged} {Locomotion}},
	volume = {7},
	issn = {2377-3766},
	doi = {10.1109/LRA.2022.3151396},
	number = {2},
	urldate = {2026-01-08},
	journal = {IEEE Robotics and Automation Letters},
	author = {Ji, Gwanghyeon and Mun, Juhyeok and Kim, Hyeongjun and Hwangbo, Jemin},
	month = apr,
	year = {2022},
	pages = {4630--4637},
}

@inproceedings{rudin_learning_2021,
	title = {Learning to {Walk} in {Minutes} {Using} {Massively} {Parallel} {Deep} {Reinforcement} {Learning}},
	language = {en},
	urldate = {2026-01-08},
	booktitle = {5th Annual Conference on Robot Learning},
	author = {Rudin, Nikita and Hoeller, David and Reist, Philipp and Hutter, Marco},
	month = jun,
	year = {2021},
}

@misc{nahrendra_dreamwaq_2023,
	title = {{DreamWaQ}: {Learning} {Robust} {Quadrupedal} {Locomotion} {With} {Implicit} {Terrain} {Imagination} via {Deep} {Reinforcement} {Learning}},
	shorttitle = {{DreamWaQ}},
	doi = {10.48550/arXiv.2301.10602},
	urldate = {2025-12-25},
	publisher = {arXiv},
	author = {Nahrendra, I. Made Aswin and Yu, Byeongho and Myung, Hyun},
	month = mar,
	year = {2023},
}

@misc{zhi_learning_2025,
	title = {Learning a {Unified} {Policy} for {Position} and {Force} {Control} in {Legged} {Loco}-{Manipulation}},
	language = {en},
	urldate = {2025-12-08},
	journal = {arXiv.org},
	author = {Zhi, Peiyuan and Li, Peiyang and Yin, Jianqin and Jia, Baoxiong and Huang, Siyuan},
	month = may,
	year = {2025},
}

@misc{gao_neural_2025,
	title = {Neural {Internal} {Model} {Control}: {Learning} a {Robust} {Control} {Policy} via {Predictive} {Error} {Feedback}},
	shorttitle = {Neural {Internal} {Model} {Control}},
	doi = {10.48550/arXiv.2411.13079},
	urldate = {2025-08-26},
	publisher = {arXiv},
	author = {Gao, Feng and Yu, Chao and Wang, Yu and Wu, Yi},
	month = may,
	year = {2025},
}

@inproceedings{huang_datt_2023,
	title = {{DATT}: {Deep} {Adaptive} {Trajectory} {Tracking} for {Quadrotor} {Control}},
	shorttitle = {{DATT}},
	language = {en},
	urldate = {2025-02-09},
	booktitle = {8th Annual Conference on Robot Learning},
	author = {Huang, Kevin and Rana, Rwik and Spitzer, Alexander and Shi, Guanya and Boots, Byron},
	month = aug,
	year = {2023},
}

@misc{kumar_rma_2021,
	title = {{RMA}: {Rapid} {Motor} {Adaptation} for {Legged} {Robots}},
	shorttitle = {{RMA}},
	doi = {10.48550/arXiv.2107.04034},
	urldate = {2025-05-02},
	publisher = {arXiv},
	author = {Kumar, Ashish and Fu, Zipeng and Pathak, Deepak and Malik, Jitendra},
	month = jul,
	year = {2021},
}

@misc{xiao_safe_2024,
	title = {Safe {Deep} {Policy} {Adaptation}},
	doi = {10.48550/arXiv.2310.08602},
	urldate = {2025-02-11},
	publisher = {arXiv},
	author = {Xiao, Wenli and He, Tairan and Dolan, John and Shi, Guanya},
	month = apr,
	year = {2024},
}

@inproceedings{lyu_rl2ac_2024,
	title = {{RL2AC}: {Reinforcement} {Learning}-based {Rapid} {Online} {Adaptive} {Control} for {Legged} {Robot} {Robust} {Locomotion}},
	isbn = {979-8-9902848-0-7},
	shorttitle = {{RL2AC}},
	doi = {10.15607/RSS.2024.XX.060},
	language = {en},
	urldate = {2026-04-21},
	booktitle = {Robotics: {Science} and {Systems} {XX}},
	publisher = {Robotics: Science and Systems Foundation},
	author = {Lyu, Shangke and Lang, Xin and Zhao, Han and Zhang, Hongyin and Ding, Pengxiang and Wang, Donglin},
	month = jul,
	year = {2024},
}

@inproceedings{long_learning_2024,
	title = {Learning {H}-{Infinity} {Locomotion} {Control}},
	language = {en},
	booktitle = {8th Annual Conference on Robot Learning},
	year={2024},
	author = {Long, Junfeng and Yu, Wenye and Li, Quanyi and Wang, Zirui and Lin, Dahua and Pang, Jiangmiao},
}

@misc{he_agile_2024,
	title = {Agile {But} {Safe}: {Learning} {Collision}-{Free} {High}-{Speed} {Legged} {Locomotion}},
	shorttitle = {Agile {But} {Safe}},
	doi = {10.48550/arXiv.2401.17583},
	urldate = {2026-04-21},
	publisher = {arXiv},
	author = {He, Tairan and Zhang, Chong and Xiao, Wenli and He, Guanqi and Liu, Changliu and Shi, Guanya},
	month = may,
	year = {2024},
}

@article{concur_learn,
  author={Kamalapurkar, Rushikesh and Reish, Benjamin and Chowdhary, Girish and Dixon, Warren E.},
  journal={IEEE Transactions on Automatic Control}, 
  title={Concurrent Learning for Parameter Estimation Using Dynamic State-Derivative Estimators}, 
  year={2017},
  volume={62},
  number={7},
  pages={3594-3601}
  }

@article{haddadin_robot_2017,
  author  = {Haddadin, Sami and De Luca, Alessandro and Albu-Sch{\"a}ffer, Alin},
  title   = {Robot Collisions: A Survey on Detection, Isolation, and Identification},
  journal = {IEEE Transactions on Robotics},
  volume  = {33},
  number  = {6},
  pages   = {1292--1312},
  year    = {2017},
  doi     = {10.1109/TRO.2017.2723903}
}

@inproceedings{magrini_control_2015,
  author    = {Magrini, Emanuele and Flacco, Fabrizio and De Luca, Alessandro},
  title     = {Control of Generalized Contact Motion and Force in Physical Human--Robot Interaction},
  booktitle = {2015 IEEE International Conference on Robotics and Automation (ICRA)},
  pages     = {2298--2304},
  year      = {2015},
  publisher = {IEEE},
  doi       = {10.1109/ICRA.2015.7139504}
}

@inproceedings{lim_proprioceptive_2023,
  author    = {Lim, Daegyu and Kim, Myeong-Ju and Cha, Junhyeok and Kim, Donghyeon and Park, Jaeheung},
  title     = {Proprioceptive External Torque Learning for Floating Base Robot and Its Applications to Humanoid Locomotion},
  booktitle = {2023 IEEE/RSJ International Conference on Intelligent Robots and Systems (IROS)},
  pages     = {8510--8517},
  year      = {2023},
  publisher = {IEEE},
  doi       = {10.1109/IROS55552.2023.10342530}
}

@article{lim_mobnet_2025,
  author  = {Lim, Daegyu and Kim, Myeong-Ju and Cha, Junhyeok and Park, Jaeheung},
  title   = {{MOB-Net}: Limb-Modularized Uncertainty Torque Learning of Humanoids for Sensorless External Torque Estimation},
  journal = {The International Journal of Robotics Research},
  volume  = {44},
  number  = {1},
  pages   = {96--128},
  year    = {2025},
  doi     = {10.1177/02783649241260428},
  note    = {First published online in 2024}
}

@inproceedings{sorrentino_ukf_2024,
  author    = {Sorrentino, Ines and Romualdi, Giulio and Pucci, Daniele},
  title     = {{UKF}-Based Sensor Fusion for Joint-Torque Sensorless Humanoid Robots},
  booktitle = {2024 IEEE International Conference on Robotics and Automation (ICRA)},
  pages     = {13150--13156},
  year      = {2024},
  publisher = {IEEE},
  doi       = {10.1109/ICRA57147.2024.10610951}
}

@article{chenDisturbanceObserverBased2004,
	title = {Disturbance observer based control for nonlinear systems},
	volume = {9},
	issn = {1941-014X},
	doi = {10.1109/TMECH.2004.839034},
	number = {4},
	urldate = {2024-05-07},
	journal = {IEEE/ASME Transactions on Mechatronics},
	author = {Chen, Wen-Hua},
	month = dec,
	year = {2004},
	pages = {706--710}}

@article{chenNonlinearDisturbanceObserver2000,
	title = {A nonlinear disturbance observer for robotic manipulators},
	volume = {47},
	issn = {1557-9948},
	doi = {10.1109/41.857974},
	number = {4},
	urldate = {2024-05-07},
	journal = {IEEE Transactions on Industrial Electronics},
	author = {Chen, Wen-Hua and Ballance, D.J. and Gawthrop, P.J. and O'Reilly, J.},
	month = aug,
	year = {2000},
	pages = {932--938}}

@inproceedings{de_luca_actuator_2003,
	title = {Actuator failure detection and isolation using generalized momenta},
	volume = {1},
	issn = {1050-4729},
	doi = {10.1109/ROBOT.2003.1241665},
	urldate = {2026-08-12},
	booktitle = {2003 {IEEE} {International} {Conference} on {Robotics} and {Automation} ({Cat}. {No}.{03CH37422})},
	author = {De Luca, Alessandro and Mattone, Raffaella},
	month = sep,
	year = {2003},
	pages = {634--639 vol.1}}

@inproceedings{de_luca_collision_2005,
  author    = {De Luca, Alessandro and Mattone, Raffaella},
  title     = {Sensorless Robot Collision Detection and Hybrid Force/Motion Control},
  booktitle = {Proceedings of the 2005 IEEE International Conference on Robotics and Automation},
  pages     = {999--1004},
  year      = {2005},
  publisher = {IEEE},
  doi       = {10.1109/ROBOT.2005.1570247}
}

@ARTICLE{robotcollision2017,
  author={Haddadin, Sami and De Luca, Alessandro and Albu-Schäffer, Alin},
  journal={IEEE Transactions on Robotics}, 
  title={Robot Collisions: A Survey on Detection, Isolation, and Identification}, 
  year={2017},
  volume={33},
  number={6},
  pages={1292-1312},
  doi={10.1109/TRO.2017.2723903}}

@misc{fan-ziqi2024rl_sar,
  author = {Fan, Ziqi},
  title = {{rl\_sar}: Simulation Verification and Physical Deployment of Robot Reinforcement Learning Algorithm},
  year = {2024}
}

@inproceedings{carpentier2019pinocchio,
   title={The Pinocchio C++ library -- A fast and flexible implementation of rigid body dynamics algorithms and their analytical derivatives},
   author={Carpentier, Justin and Saurel, Guilhem and Buondonno, Gabriele and Mirabel, Joseph and Lamiraux, Florent and Stasse, Olivier and Mansard, Nicolas},
   booktitle={IEEE International Symposium on System Integrations (SII)},
   year={2019}
}

@misc{rudin2021leggedgym,
  title={Isaac Gym Environments for Legged Robots},
  author={Rudin, Nikita},
  year={2021},
}

@article{makoviychuk2021isaac,
  title={Isaac gym: High performance gpu-based physics simulation for robot learning},
  author={Makoviychuk, Viktor and Wawrzyniak, Lukasz and Guo, Yunrong and Lu, Michelle and Storey, Kier and Macklin, Miles and Hoeller, David and Rudin, Nikita and Allshire, Arthur and Handa, Ankur and State, Gavriel},
  journal={arXiv preprint arXiv:2108.10470},
  year={2021}
}

@ARTICLE{DeepDOB_locomotion,
  author={Muhamad, Fikih and Kusuma, Anak Agung Krisna Ananda and Park, Jae-Han and Kim, Jung-Su},
  journal={IEEE Robotics and Automation Letters}, 
  title={Enhancing Robustness of Locomotion Policy for Quadrupedal Robot With Deep Disturbance Observer}, 
  year={2025},
  volume={10},
  number={9},
  pages={9376-9383},
  doi={10.1109/LRA.2025.3595037}}

@misc{lyu2026ADAPT,
      title={ADAPT: Analytical Disturbance-Aware Policy Training for Humanoid Locomotion}, 
      author={Bofan Lyu and Jindou Jia and Kuangji Zuo and Yanshuo Lu and Shijia Han and Gen Li and Boyu Ma and Jingliang Li and Geng Li and Jianfei Yang},
      year={2026},
      eprint={2606.16542},
      archivePrefix={arXiv},
      primaryClass={cs.RO},
}

\end{document}